\documentclass[11pt,a4paper]{article}
\usepackage[hyperref]{acl2021}
\usepackage{times}
\usepackage{latexsym}
\usepackage{xspace}
\usepackage{CJK}
\usepackage{makecell}
\usepackage{booktabs}
\usepackage{siunitx}
\usepackage{bm}
\usepackage[utf8]{inputenc}
\usepackage{graphicx}
\usepackage[english]{babel}
\babelprovide[import]{thai}
\babelprovide[import]{hindi}
\babelprovide[import]{sanskrit}

\usepackage{microtype}
\usepackage{multirow}
\usepackage{paralist}

\newcommand{\eg}{\emph{e.g.},\xspace}
\newcommand{\ie}{\emph{i.e.},\xspace}

\aclfinalcopy % Uncomment this line for the final submission
\title{As Easy as 1, 2, 3: Behavioural Testing of NMT Systems \\for Numerical Translation}

\author{Jun Wang,$^1$ Chang Xu,$^1$ Francisco Guzm\'an,$^2$ Ahmed El-Kishky,$^3$\thanks{~~This work was conducted while author was working at Facebook AI}\\
    \bf Benjamin I. P. Rubinstein,$^1$ Trevor Cohn$^1$ \\
  $^1$University of Melbourne, Australia \\
  $^2$Facebook AI, $^3$Twitter Cortex\\
  \texttt{jun2@student.unimelb.edu.au}\\
  \texttt{\{xu.c3,benjamin.rubinstein,trevor.cohn\}@unimelb.edu.au}\\
  \texttt{fguzman@fb.com, aelkishky@twitter.com}
  }

\date{}

\begin{document}
\maketitle
\begin{abstract}

Mistranslated numbers have the potential to cause serious effects, such as financial loss or medical misinformation.
In this work we develop comprehensive assessments of the robustness of neural machine translation systems to numerical text via behavioural testing.
We explore a variety of numerical translation capabilities a system is expected to exhibit and design effective test examples to expose system underperformance.
We find that numerical mistranslation is a general issue: major commercial systems and state-of-the-art research models fail on many of our test examples, for high- and low-resource languages.
Our tests reveal novel errors that have not previously been reported in NMT systems, to the best of our knowledge.
Lastly, we discuss strategies to mitigate numerical mistranslation.%
%\footnote{Data and code to be made available upon acceptance.}
\end{abstract}

\section{Introduction}
Just as neural machine translation (NMT) systems have achieved tremendous benchmark results, they have been proven brittle when faced with irregular inputs such as noisy text~\cite{belinkov2017synthetic,michel-neubig-2018-mtnt} or adversarial inputs~\cite{cheng-etal-2020-advaug}.
Among such errors, mistranslation of \textit{numerical text} constitutes a crucial but under-explored category that may have profound implications.
For example, in the medical domain, mistranslating the number of confirmed cases of a contagious disease like COVID-19 may exacerbate public health misinformation. %received by populations using low-resourced languages (\eg Tamil).
Numerical errors made in  financial document translation, \eg an extra or omitted digit or decimal point, could lead to  significant monetary loss.
Surprisingly, we find that numerical mistranslation is a general issue faced by state-of-the-art NMT systems, including commercial and research systems, with evidence present across  contexts: for both high and low resource languages, and for both close and distant languages.

% \begin{table}[t!]
%     \centering
%     %\small
%     \footnotesize
%     %\scriptsize
%     \begin{tabular}{@{}p{1.2cm}p{2.6cm}p{2.7cm}}
%       \toprule
%       \multicolumn{1}{c}{\textbf{Type}} & \multicolumn{1}{c}{\textbf{Input}} & \multicolumn{1}{c}{\textbf{Output}} \\
%       \midrule
%       Separators & The distance is 557,601.101 meters. & \textbf{[De]}: Die Entfernung beträgt 557.601\textcolor{red}{.}101 Meter.\\
%       \hline
%       Numerals & The total weight is \textcolor{red}{two hundred and two} kg. & \textbf{[Zh]}:总重量为\textcolor{red}{220}公斤。\\
%       \hline
%       Digits & The R0 of the disease is \textcolor{red}{3.28}. & \textbf{[Ne]}: \texthindi{रोगको R0 \textcolor{red}{28.२28} हो।} \\
%       \hline
%       Units  & There were \textcolor{red}{100.01 million} cases worldwide. & \textbf{[Zh]}:全世界有\textcolor{red}{1.001亿}病例。(100.1 million)\\
%       \bottomrule
%     \end{tabular}
%     \caption{Numerical errors discovered by our method when behavioural testing two popular commercial translation systems using their public APIs.}
%     \label{tab:examples}
% \end{table}

\begin{table}[]
    \footnotesize
    \centering
    \includegraphics{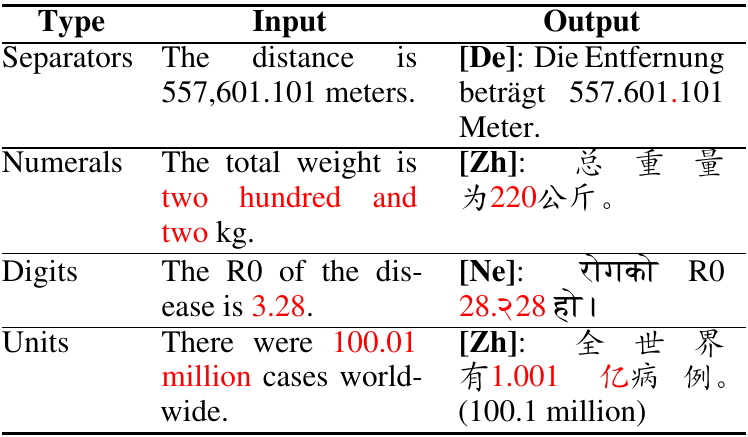}
    \caption{Numerical errors discovered by our method when behavioural testing two popular commercial translation systems using their public APIs.}
    \label{tab:examples}
\end{table}

De facto standard metrics such as BLEU~\cite{papineni-etal-2002-bleu} may fail to flag a numerical translation error, which only contributes a very minor penalty, as it is typically a single-token mistranslation.
%\chang{Discussing some traditional evaluation metric for MT (\eg BLEU) that may fail to flag  TC: Yeah, effectively it's just a single token, so it gets a very minor penalty. There's a history of BLEU missing things like this, the same is true for transliteration of names.}
To facilitate the discovery of numerical errors made by NMT systems, we propose a black-box test method\footnote{Our code is available  at \url{https://github.com/JunW15/NumberTest}} for assessing and debugging the numerical translation of NMT systems in a systematic manner.
Our method extends the \texttt{CheckList} behavioural testing framework~\cite{ribeiro-etal-2020-beyond} by designing automatic test cases to assess a suite of fundamental capabilities a system should exhibit in translating numbers.

Our tests on state-of-the-art NMT systems expose novel error types that have evaded close examination (Table~\ref{tab:examples}).
These error types greatly extend the \emph{number} category (\texttt{NUM}) of the \emph{catastrophic errors}~\cite{specia-etal-2020-findings} of NMT systems with richer error types.
Finally, the abuse of these errors constitute vectors of attack: error-prone numerical tokens injected into monolingual data may corrupt \emph{back-translation-based} training, as the resulting back-translated sentences are very likely to contain the desired errors.

\section{Method}

We follow \citet{ribeiro-etal-2020-beyond}'s \texttt{CheckList} in designing our evaluation suite for NMT systems: 
we present several basic \emph{capabilities} an NMT system should be expected to exhibit in translating common everyday numerical text; 
we then generate \emph{test examples} specific to each capability to benchmark performance and find bugs in NMT systems. 

\subsection{Capabilities of Translating Numbers}
We explore four capabilities (see Table~\ref{tab:capabilities}), 
demonstrating expected translation ability of a system on common types of numerical text.
Concretely, the \emph{Integers} and \emph{Decimals} represent basic capabilities;  
they can be manifested by testing on sequences of digits with variable lengths (\eg 100 vs. 10000) or decimals with the decimal mark placed at varying locations (1.001 vs. 10.01).
We find that the tested NMT systems are more likely to malfunction when translating larger integers and decimals with longer fractional parts.
The \emph{Numerals} capability pertains to whether a system is able to translate numbers that are presented as words.
%For \emph{Formats}, we test whether a system can handle formatted text comprised of numbers and special symbols (\eg \$, /, °C, etc.).
The \emph{Separators} capability checks if a model can deal with numbers containing decimal or thousands separators.%
\footnote{While \emph{Decimals} and \emph{Separators} may have overlapping instances (\eg the decimal mark), their specific formats in our testing are different (Table~\ref{tab:capabilities}), which leads us to find non-overlapping error types: most \emph{Decimals} errors involve translating numbers into wrong digits, whereas \emph{Separators} errors pertain to mistranslation in localisation usage (\eg German and English use different decimal and thousand separators).}
%Lastly, the \emph{Measurements} capability inspects if a system can translate numbers accompanied by units of measurement (\eg metre).
%To exhibit this capability, a system has to understand the units of measurement used in each language, and perform necessary conversions if the units do not match between the languages.
Systems that fail to manifest one or more of these capabilities may produce wrong numbers that can be inconspicuous to users and become a ready, exploitable source of misinformation.

\begin{table}[t!]
\footnotesize
    \centering
    \begin{tabular}{p{1.6cm}p{5cm}}
    \toprule
    \textbf{Capability} & \multicolumn{1}{c}{\textbf{Examples}}\\
    %\hline
    \midrule
    Integers & There were \textcolor{red}{914} confirmed cases of COVID-19 reported yesterday.\\
    %\hline
    \midrule
    Decimals & The reproduction number of COVID-19 is between \textcolor{red}{3.28} and \textcolor{red}{5.70}. \\
    %\hline
    \midrule
    Numerals & The total amount of transfer is \textcolor{red}{fifty-two dollars and seven cents}.\\
    %\hline
    \midrule
    %Formats & \$8 (currency), 08/08/2021 (date), 8pm (time), 8\% (percentage), 8°C (temperature), -8 (negative)\\
    Separators & 123\textcolor{red}{,}456\textcolor{red}{.}12 (En) $\rightarrow$ 123\textcolor{red}{.}456\textcolor{red}{,}12 (De)\\
    %\hline
    %Measurements & The total length is \textcolor{red}{116 meters}.\\
    %\hline
    \bottomrule
    \end{tabular}
    \caption{Tested capabilities of NMT systems in translating common types of numerical text.}
    \label{tab:capabilities}
\end{table}

\subsection{Test Examples}
To efficiently test the identified capabilities across multiple systems on distinct language pairs, we generate desired test examples using templates.
For example, to test the \emph{Numerals} capability, we use a template sentence such as ``CNBC reported there were at least [\texttt{NUM}] cases worldwide.'', where ``[\texttt{NUM}]'' is a number with the format ``\texttt{ddd.ddn}'', consisting of multiple digits and a numeral (\eg 100.01 million).
%\newadd{Most of the test templates come from medicine and finance is due to our expectation that numerical mistranslations in these domains would have more profound implications than elsewhere (\eg{news}). We speculate that these errors may also present in pretty much any context. Appendix~\ref{transfer} shows how to transfer our test cases to another domain.}

We experiment with formats of various lengths and decimal-point positions.
We fill a format with random digits and numerals, and explore 25 different formats across all capabilities.
This allows us to generate a diversity of numbers at scale, akin to fuzzing a program with random inputs to uncover bugs. 
We also note that all the numbers created for a format can be seen as a set of ``adversarial'' examples, as they are small perturbations of each other.
%When translating between English and a language using a different scale (\eg Chinese numbers group by ten-thousands), an error may occur if a system fails to translate the scale units (\eg 10 thousand$\rightarrow$1万).
Details about the test examples for each capability and the testing process can be found in Supplementary material.

\section{Evaluation}
\label{sec:eval}

%\subsection{Setup} % commenting out since there's no leading text above this heading.
%\label{sec:setup}

Before presenting experimental results and discussion of our test framework, we first detail our evaluation setup.

\paragraph{Language pairs.}
We test both high-resource (HR) and low-resource (LR) scenarios.
For HR, we consider two language pairs: English-German and English-Chinese, and for LR, we focus on English-Tamil and English-Nepali.
We test both translation directions for each pair.

\paragraph{SOTA systems.}
We conduct behavioural testing against two popular commercial translation systems (denoted by \textbf{A} and \textbf{B}).
%\footnote{\newadd{While the online systems change with time, they do represent the current best systems available. Added to ``small matters'' that do not show up in BLEU score based evaluation is often better handled by commercial systems.}} 
As research systems, we use pre-trained models that were shown to perform well in WMT competitions (denoted by \textbf{R}),
specifically, fairseq's transformer for English-German~\cite{ng-etal-2019-facebook}, English-Tamil~\cite{chen-etal-2020-facebook}, and English-Chinese/Nepali~\cite{fomicheva2020unsupervised}.

\paragraph{The evaluation metric.}
For each capability we curate a list of test examples (sentences containing numbers), which are taken from various sources, including existing corpora or manually crafted (details in Supplementary material).
To these sentences we remove the number component, and replace it with a number based on the specific capability being tested.
This test collection is then input to a translation system, and we report the \emph{Pass Rate} (PR), the fraction of inputs where the system translates the numerical component perfectly.\footnote{We count a Pass if the output matches the ground truth number, allowing for the use of digits (Arabic or local scripts) or numerals. For this purpose we use \texttt{num2words} (\url{https://pypi.org/project/num2words/}), \texttt{cn2an} (\url{https://github.com/Ailln/Cn2An.jl}), and locally developed scripts (for Ne and Ta).} 
%We report the performance separately for each capability. %, and list of test examples (sentences containing numbers) for each capability.
%The PR is then defined as the proportion of passed examples on that capability.
%A \emph{Pass} is counted if the system correctly translates the number in an example.
%To detect a \emph{Pass}, we compare the numbers in the output to the ground-truth.
%For numeral outputs, we use tools like \texttt{text2digits} (\texttt{cn2an} for Chinese) to convert them into digits before comparing to the ground-truth.
%On low-resource languages (Ne and Ta) for which such text-to-digit tools are unavailable, we develop scripts for the conversion ourselves.

\subsection{Testing Performance}

Table~\ref{tab:results} shows the results of testing the three SOTA systems across the HR/LR language pairs.

Among the four capabilities,  \emph{Numerals} turns out to be the most challenging  across the systems tested, with the average $\overline{\textrm{PR}}$ of 70.8\%.
This is probably because, compared to other forms, numbers are less frequently written as words, resulting in insufficient examples available for training.
At the other extreme,  \emph{Integers}, which tests on pure digits, is the easiest capability, as expected.
Despite this, it is not a `solved problem', given all systems report imperfect \textrm{PR}~$<100$ on at least one language.

Across the systems, the research system \textbf{R} ($\overline{\textrm{PR}}$: 77.8\%) underperforms the two commercial ones ($\overline{\textrm{PR}}_A$: 80.6\%, $\overline{\textrm{PR}}_B$: 90.4\%).
This is largely caused by the fact that the research system fails markedly on the En$\rightarrow$Ne direction.

Per language, the results are similar in both translation directions, implying that numerical translation is a symmetric problem.
Note that the results on LR are not always worse than that on HR (PRs on En-Ta are surprisingly the highest of all).
This  suggests that the size of training data is not the sole factor for  high-quality numerical translation.

\begin{table*}[t!]
% \small
\centering
\footnotesize
\def\arraystretch{1.1}
\sisetup{
table-figures-integer = 3,
table-figures-decimal = 1,
table-number-alignment = center,
%table-parse-only,
detect-weight=true,
detect-inline-weight=text
}
\robustify\bfseries% see 7.15 in siunitx.pdf
\newcommand{\tbnum}[1]{\bfseries #1}
%\begin{tabular}{c|c|c|c|c|c|c|c|c|c|c|c|c|c}
%\begin{tabular}{c|SSS|SSS|SSS|SSS|S}
\begin{tabular}{p{6ex}SSSSSSSSSSSSS@{}}
\toprule
\multirow{2}{*}{\textbf{Lang}}   & \multicolumn{3}{c}{\textbf{Integers}} & \multicolumn{3}{c}{\textbf{Decimals}} & \multicolumn{3}{c}{\textbf{Numerals}} & \multicolumn{3}{c}{\textbf{Separators}} & {\multirow{2}{*}{\textbf{Avg}}} \\ 
\cmidrule(lr){2-4}
\cmidrule(lr){5-7}
\cmidrule(lr){8-10}
\cmidrule(lr){11-13}
%\textbf{System} 
& {\textbf{A}}  & {\textbf{B}}  & {\textbf{R}} &  {\textbf{A}}  & {\textbf{B}}  & {\textbf{R}} &  {\textbf{A}}  & {\textbf{B}}  & {\textbf{R}} &  {\textbf{A}}  & {\textbf{B}}  & {\textbf{R}} &  \\  
\toprule
\textbf{En$\rightarrow$Zh}  & 100.0       & 100.0       & 94.0           & 100.0       & 100.0       & 92.2       & 77.5        & 72.5        & 67.5      & 100.0        & 100.0       & 91.4     & 91.2   \\
\textbf{Zh$\rightarrow$En}  & 94.0        & 78.0        & 90.0           & 100.0       & 100.0       & 93.8       & 82.0        & 78.0        & 56.7      & 100.0        & 100.0       & 83.3     & 88.0  \\ \midrule
\textbf{En$\rightarrow$De}  & 100.0       & 100.0       & 100.0          & 93.8        & 78.1        & 68.8       & 87.5        & 67.5        & 95.0      & 83.3         & 80.0        & 80.0     & 86.2    \\ 
\textbf{De$\rightarrow$En}  & 100.0       & 100.0       & 100.0          & 98.4        & 79.7        & 95.3       & 87.0        & 84.0        & 65.7      & 97.1         & 68.5        & 65.7     & 86.8    \\ \midrule
\textbf{En$\rightarrow$Ta}  & 100.0       & 98.0        & 100.0          & 100.0       & 100.0       & 100.0      & 100.0       & 97.5        & 100.0     & 100.0        & 100.0       & 100.0    & 99.6    \\   
\textbf{Ta$\rightarrow$En}  & 98.0        & 100.0       & 100.0          & 98.4        & 100.0       & 98.4       & 96.0        & 100.0         & 90.0      & 100.0        & 88.6        & 100.0    & 97.4    \\ \midrule
\textbf{En$\rightarrow$Ne}  & 18.0        & {-}           &        70.0        & 17.2        & {-}           &     72.0           & 7.5         & {-}           &       65.0         & 16.7         & {-}           &      60.0    &  \tbnum{40.8}   \\ 
\textbf{Ne$\rightarrow$En}  & 98.0        & {-}           &    88.0        & 96.9        & {-}           &        1.6     & 46.7        & {-}           &     5.3        & 80.0           & {-}           &      0.0   & \tbnum{52.1}   \\ 
\bottomrule
\textbf{Avg}    &   88.5      &     96.0    &     92.8           & 88.1        & 92.9        &    77.8            & \tbnum{61.1}        & 83.2        &     \tbnum{68.2}           & 84.6         & 89.5        &     \tbnum{72.6}     & {-}   \\ 
\bottomrule
\end{tabular}
\caption{Test results (Pass Rate \%) on the capabilities for numerical translation, with low averaged scores in bold. Nepali is not supported by System \textbf{B}.}
\label{tab:results}
\end{table*}

\subsection{Error Analysis}
We present analysis of novel types of mistranslations discovered from testing.

\begin{table*}[t!]
    \footnotesize
    \centering
    \includegraphics{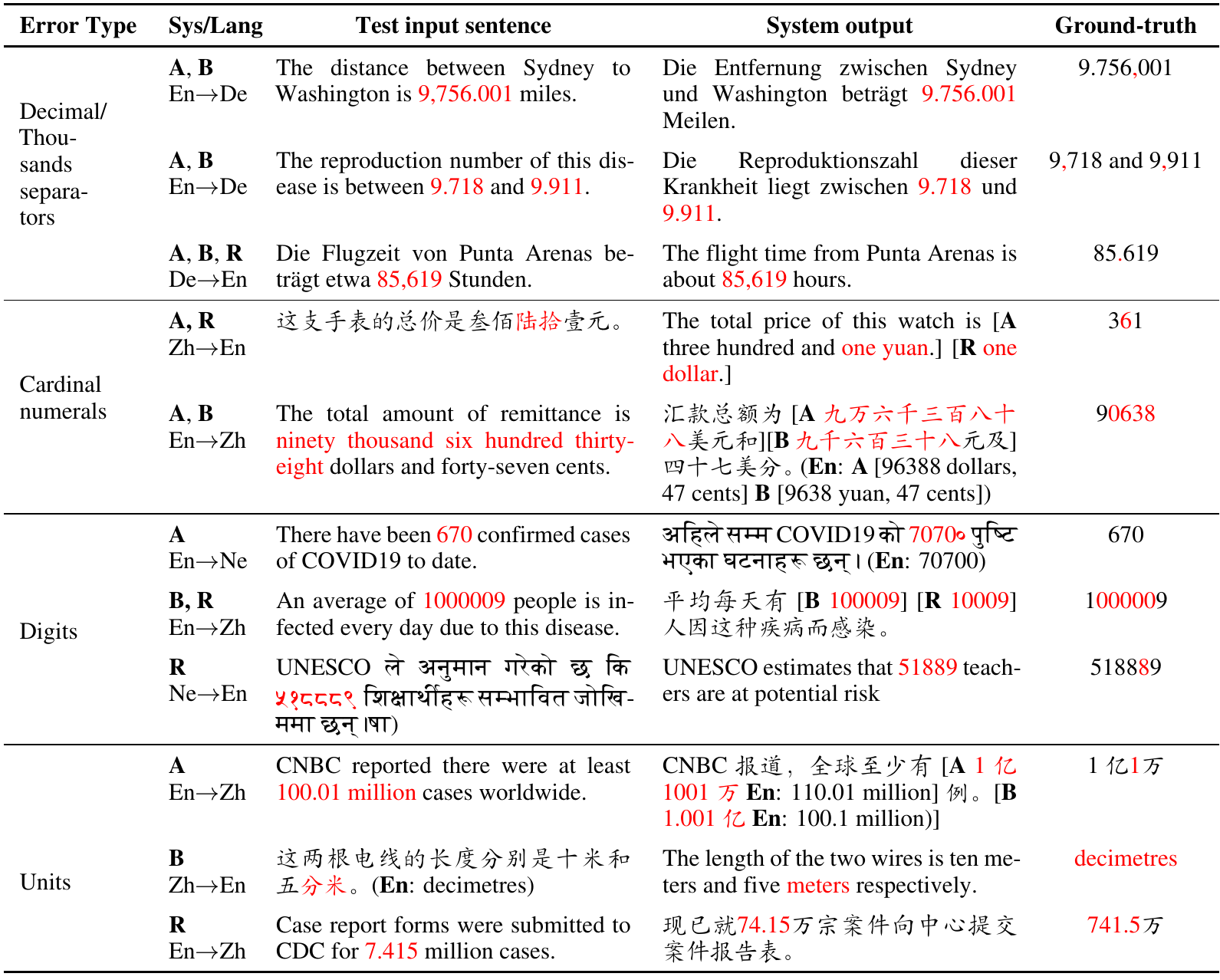}
    \caption{Examples of four major types of errors discovered by our tests on three SOTA NMT systems.}
    \label{tab:errors}
\end{table*}

\paragraph{Decimal/thousands separators.}
We find that the decimal/thousands separators are prone to be mistranslated in localisation scenarios, when conventions differ between the languages (\eg ``,'' and ``.'' are the thousands and decimal separators in English while they are swapped in German).
A common type of error is that a separator remains the same after translation (Table~\ref{tab:errors}, row 1).
This is probably due to the lack of sufficient training data to learn the translation of the separators in the target language.
%The most common mistake in digital translation between English and German is the misuse of separators. Translating 1,000[en] into 1,000[de] cause the error of 999. When translating money figures, or some important statistics, this kind of error can cause serious consequences.

\paragraph{Cardinal numerals.}
Cardinal numerals are commonly used in commercial and financial contexts.
For example, the \emph{financial characters} (\eg \begin{CJK*}{UTF8}{gbsn}``壹''\end{CJK*} meaning one) are typical in Chinese financial documents.
However, we find that the tested translation systems perform fairly poorly in translating cardinal numerals (Table~\ref{tab:errors}, row 2). 
Common errors include mistranslation or under-translation of the unit words (\eg hundred) or the number words (\eg \begin{CJK*}{UTF8}{gbsn}``陆拾''\end{CJK*}).
Most often, the errors appear to be caused by the unique unit words used in different languages (\eg \begin{CJK*}{UTF8}{gbsn}``万''\end{CJK*} in Chinese equals to 10 thousand), where a system needs to ``compute'' the correct amount for translation.

\paragraph{Digits.}
The pure digit translation (10$\rightarrow$10) is expected to be easy, since a system may opt to copy the entire number as the translation. 
However, we find that the digit translation between English and low-resource languages can be far from satisfactory.
An example is the translation between English and Nepali (Table~\ref{tab:errors}, row 3).
One reason for this result is that Nepali has its own numerals for digits. 
As a result, a system would try to convert a digit into a Nepali digit (instead of keeping it unchanged) when translating numbers, which is difficult given limited training resources~\cite{guzman-etal-2019-flores}.
Another common issue in digit translation is handling repeats of the same digit.
A system is prone to omit or add one or more digits in the translation.

\paragraph{Units.}
This error often occurs when translating numbers accompanied by units of measurements (\eg 10 meters), especially when the target unit is unique to the language, \eg \begin{CJK*}{UTF8}{gbsn}``角''\end{CJK*} in Chinese means ``10 cents''.
In such cases (Table~\ref{tab:errors}, last row), the system may need to learn the implicit conversion rules and then use them to ``calculate'' the correct numbers with the target unit of measurement.
For example, when translating ``10.01 million'' into ``1001\begin{CJK*}{UTF8}{gbsn}万''\end{CJK*} in Chinese, the system has to convert ``10.01'' into ``1001'' and then use the correct unit \begin{CJK*}{UTF8}{gbsn}``万''\end{CJK*}.
An error may occur if the system fails either or both stages of this process (\ie mistranslating the numbers and/or units).

%\subsection{Errors as Adversarial Examples}

\section{Potential Mitigation Strategies}
Finally, we discuss several strategies that may mitigate the above errors discovered by our method\footnote{We leave validation of these strategies to future work.}.
%\newadd{This paper focused on the test method for behavioural testing for numerical translation in NMT systems. We leave how to solve it as an open question. However, we will discuss several strategies to mitigate the above errors discovered by our framework in this section and will validate them in future work.}

\paragraph{Separate treatment of numbers.}
Although NMT models have been shown capable of performing basic arithmetic or bracket matching~\cite{suzgun-etal-2019-lstm}, this paper demonstrates that handling the various forms of numerical text in reality is still challenging.
It may be worth separating numerical translation out into an individual process, as in Statistical MT~\cite{koehn2009statistical}, that identifies numbers in the input, applies specific translation rules to them, and incorporates the translation into the output~\cite{tu2012universal}.

\paragraph{Data augmentation.}
Training with more quality data leads to better translation quality~\cite{barrault-etal-2020-findings}.
In our testing, we observe a large proportion of errors (\eg financial characters, units) stemming from mistranslation of specific numerals that are unique or used less frequently (\eg \begin{CJK*}{UTF8}{gbsn}``角''\end{CJK*}, decimetres) in a language. 
Such errors could potentially be reduced if more \emph{numeral-specific} instances were added to training.

\paragraph{Tailoring BPE segmentation.}
The Byte Pair Encoding (BPE) has been used by most leading NMT systems.
However, long sequences of digits or numbers with separators (\eg ``,'', ``.'') are often split into varying sized fragments by BPE.
This would render learning more difficult, as the system has to account for the dependency between the partitions.
To circumvent this, one may wish to segment numbers differently, \eg to encode all numbers as character sequences, or as meaningful groupings of components (\eg segment into groups of 3 digits when processing English.)

\paragraph{Sanity checks.}
It is helpful to post-check whether all numbers in a translation are correct by comparing them to the inputs.
This could be automated in the same way as we measure the Pass Rate (\S\ref{sec:eval}), and once again drawing parallels to software testing, could be fully automated via continuous integration of NMT systems.

\section{Conclusion}
In this paper, we propose an evaluation method to systematically assess four fundamental capabilities of NMT systems in translation numbers by virtue of a variety of test cases.
Our tests reveal novel types of errors that are general across multiple SOTA translation systems for both high and low resource languages.
We hope that our study will help improve numerical translation quality and reduce misinformation caused by numerical mistranslation.

\section*{Acknowledgements}
We thank all anonymous reviewers for their constructive comments. The authors acknowledge funding support by Facebook. %All algorithms newly described are dedicated to the public domain.

\section*{Impact Statement}

This work aims to improve the performance of NMT systems. The impact of poor numerical translation may go beyond poor user experience, potentially leading to financial loss, medical misinformation, and even a vector for poisoning NMT systems. This paper's behavioural testing could be used by an attacker to uncover flaws in a commercial NMT system. However, as in attack research in the security community, responsible highlighting of such flaws serves the purpose of improving systems: knowledge of systemic flaws in numerical translations helps vendors improve their systems to mitigate these effects in the first place, while concerted attackers are likely to discover vulnerabilities independently.

% \paragraph{Chinese-English}
% \begin{itemize}
%     \item Financial character
%     \item Wrong digital number
%     \item Large number in words
%     \item Unit conversion
%     \item ambiguous
%     \item Translating "AND"
% \end{itemize}

% \paragraph{English-German}
% \begin{itemize}
%     \item Large number in words
%     \item euro-number format
% \end{itemize}
% \paragraph{English-Nepali}
% \begin{itemize}
%     \item Wrong digital number
%     \item large number in words
% \end{itemize}

\bibliographystyle{acl_natbib}
\bibliography{reference}

% \clearpage
\appendix

\section{Appendix}

\subsection{Testing Process}

\begin{table*}
\centering
\small
\renewcommand{\arraystretch}{1.1}
\begin{tabular}{cp{9cm}p{1.2cm}p{2cm}}
\toprule
\textbf{Capability}   & \multicolumn{1}{c}{\textbf{Template}} & \textbf{Count}   & \textbf{Remarks} \\ 
\midrule
\multirow{7}{1.5cm}{\textbf{Integers}   e.g.,~5}     &   1. As of March 28, 2020, a total of [\texttt{NUM}] laboratory-confirmed COVID-19 cases (Figure) were reported to CDC.
     &  \multirow{7}{*}{50}     & \multirow{7}{2cm}{Ranging from 1 to 10 digits}       \\ 
     & 2. Case report forms were submitted to CDC for [\texttt{NUM}] cases.&&\\
&3. UNESCO estimates [\texttt{NUM}] learners are potentially at risk (pre-primary to upper-secondary education).&&\\
& 4. There have been [\texttt{NUM}] confirmed cases of COVID19 to date. && \\
& 5. CNBC reported there were at least [\texttt{NUM}] cases worldwide.
&&\\
\midrule
\multirow{5}{1.5cm}{\textbf{Decimals} e.g.,~7.14}     & 1. An average of [\texttt{NUM}] people is infected every day due to this disease &  \multirow{5}{*}{40}     & \multirow{5}{2cm}{Ranging from 1 to 4 decimal position} \\ 
& 2. The distance between Sydney to Washington is [\texttt{NUM}] miles &&\\
& 3. The genome size of the coronavirus is approximately [\texttt{NUM}] &&\\
& 4. At this point, Rosberg was about [\texttt{NUM}] seconds behind his teammate. &&\\
& 5. The reproduction number of this disease is between [\texttt{NUM}] and [\texttt{NUM}]. &&\\
\midrule
\multirow{8}{1.5cm}{\textbf{Numerals} e.g.,~five hundred}     &     1. The total amount of remittance is [\texttt{NUM}].     &  \multirow{8}{*}{40}     & \multirow{8}{2cm}{\{hundred, thousand, million, trillion\}}     \\ 
& 2. Case report forms were submitted to CDC for [\texttt{NUM}] cases.&&\\
& 3. As of 8 April 2020, approximately [\texttt{NUM}] learners have been affected due to school closures in response to COVID-19. &&\\
& 4. As of December 2019, [\texttt{NUM}] cases of MERS-CoV infection had been confirmed by laboratory tests &&\\
& 5. They then planned an ambitious open-air concert in Tokyo, with a stage costing [\texttt{NUM}] dollars US.&&\\
\midrule
\multirow{7}{1.5cm}{\textbf{Separators} e.g.,~12,230}   &   1. An average of [\texttt{NUM}] people is infected every day due to this disease       &  \multirow{7}{*}{35}     & \multirow{7}{2cm}{Ranging from 4 to 10 digits}        \\ 
& 2. The distance between Sydney to Washington is [\texttt{NUM}] miles &&\\
& 3. The genome size of the coronavirus is approximately [\texttt{NUM}] &&\\
& 4. They then planned an ambitious open-air concert in Tokyo, with a stage costing [\texttt{NUM}] dollars US. &&\\
& 5. As of December 2019, [\texttt{NUM}] cases of MERS-CoV infection had been confirmed by laboratory tests &&\\
\bottomrule
%\textbf{Measurements} &          &         &         \\ 
\end{tabular}
\caption{Summary of test examples used in our behavioural testing of NMT systems in translating numbers between English and \{German, Chinese, Nepali, Tamil\}.}
\label{tab:test-examples}
\end{table*}

Our behaviour testing proceeds in four steps.

\noindent \textbf{1) Test template selection}: we select English sentences from quality real corpora TICO-19~\cite{anastasopoulos-etal-2020-tico} and WikiMatrix~\cite{schwenk2019wikimatrix}.
TICO-19 contains documents about COVID-19 (\eg scientific articles,  conversations, Wikipedia entries) between English and 36 languages.
WikiMatrix consists of parallel sentences extracted from Wikipedia articles in 85 languages.
We randomly select five sentences with each containing numbers for the evaluation of each capability. 
We make each sentence a template by replacing the contained number with the placeholder ``[\texttt{NUM}]''.

\noindent \textbf{2) Template filling-in}: we fill the templates with randomly generated numbers in the digital format (\eg 1230000). Then, we convert the digital number into desired formats for testing (\eg 1,230,000 for a separator or 1.234 million for a numeral).

\noindent \textbf{3) Translation}: To test a system, we use it to translate all the test examples across all capabilities, and collate the translation results.

\noindent \textbf{4) Evaluation}: Finally, we check the correctness of the number translation by comparing the number to that in the input.
We account for various forms of the number (\eg for the number 5, we also consider 5.0 and five as correct translations) so as to reduce the false positives.
We also manually examine the incorrect translations to ensure they are not false positives.
%1. Find examples from  and WikiMatrix dataset, mark number as [\texttt{NUM}]
%2. Generate random numbers within the test range, convert number into the test format (Digital, Separator, Numeral ..)
%3. Translate it into target source.
%4. Only check does the number translate correct, we consider different formats of the translating number. (for example, Ground Truth: 5.0, Correct: Five, 5.0, 5)
%5. Look at the wrong cases manually for double check.
\subsection{Create (Transfer) Test Cases to a New Domain}\label{transfer}
Our testing framework facilitates constructing test instances for new domains in the following steps:
\begin{compactenum}
    \item Obtain a large corpus of text that contains numbers (\eg{CommonCrawl});
    \item Check if there is a number in the output translation;
    \item If so, then test if the output number is the correct ``translation'' for the number in the source sentence;
    \item Use instances that pass this test as templates for switching in our different numbers.
\end{compactenum}

\subsection{Test Examples Details}
Table~\ref{tab:test-examples} shows the details of the test examples used in our behavioural testing, including the templates used and the types of digits and numerals generated to fill in the templates. 

\end{document}